\documentclass[letterpaper,10pt,conference]{ieeeconf}
\IEEEoverridecommandlockouts
\usepackage{cite}
\usepackage{graphicx}
\let\labelindent\relax
\usepackage{amsmath,amssymb,enumitem}
\usepackage{hyperref}
\usepackage{booktabs}
\usepackage{xcolor}
\usepackage{caption}
\usepackage{placeins}
\usepackage{multirow}
\usepackage{bm}
\usepackage{siunitx}
\usepackage{multirow}

\definecolor{haozhipurpole}{HTML}{7A85C1}

\definecolor{darkblue}{RGB}{0,0,139}

\newcommand{\paragraphc}[1]{\vspace{0.2em}\noindent\textbf{#1}}

\definecolor{pmcolor}{RGB}{70,70,70}
\newcommand{\pmcolor}[1]{\textcolor{pmcolor}{#1}}
\newcommand{\pms}[2]{#1{\tiny{{\pmcolor{{$\pm$#2}}}}}}

\definecolor{citecolor}{HTML}{0071bc}
\hypersetup{
    colorlinks=true,
    citecolor=citecolor,
    filecolor=citecolor,
    linkcolor=citecolor,
    urlcolor=citecolor
}

\title{\LARGE \bf Learning Dexterous Manipulation Skills from Imperfect Simulations}

\author{
Elvis Hsieh$^{*}$, Wen-Han Hsieh$^{*}$, Yen-Jen Wang$^{*}$, Toru Lin, Jitendra Malik, Koushil Sreenath$^{\dagger}$, Haozhi Qi$^{\dagger}$ \vspace{0.5em}\\
UC Berkeley
\thanks{${}^*$ Equal contribution (listed in alphabetical order).}
\thanks{${}^\dagger$ Equal advising.}
}

\newcommand{\ours}{DexScrew}
\newcommand{\website}{https://dexscrew.github.io}

\begin{document}

\thispagestyle{empty}
\pagestyle{empty}

\let\oldtwocolumn\twocolumn
\renewcommand\twocolumn[1][]{%
\oldtwocolumn[{#1}{
\begin{center}
    \vspace{-1.5em}
    \includegraphics[width=\linewidth]{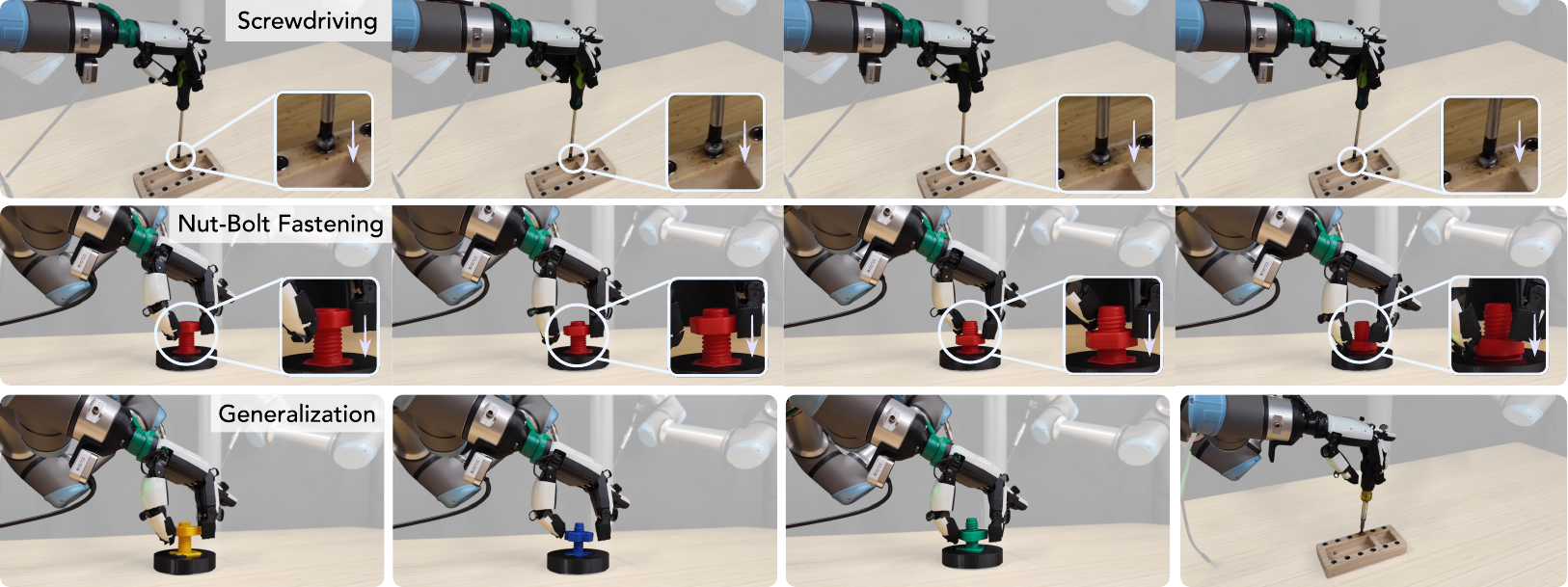}
    \captionof{figure}{We propose \ours, a sim-to-real framework for learning dexterous manipulation skills when the environment cannot be accurately simulated. In simulation, we use simplified objects to learn transferable rotational skills, which are then used to collect data and train tactile policies in the real world. We demonstrate the framework on contact-rich screwdriving (top row) and nut-bolt fastening (middle row). We also show generalization across different objects (bottom row). More videos and code are available on \href{\website}{\textcolor{citecolor}{\website}}.}
    \label{fig:teaser}
    \vspace{-0.2em}
\end{center}
}]
}

\maketitle

\begin{abstract}
Reinforcement learning and sim-to-real transfer have made significant progress in dexterous manipulation. However, progress remains limited by the difficulty of simulating complex contact dynamics and multisensory signals, especially tactile feedback. In this work, we propose \ours, a sim-to-real framework that addresses these limitations and demonstrates its effectiveness on nut-bolt fastening and screwdriving with multi-fingered hands. The framework has three stages. First, we train reinforcement learning policies in simulation using simplified object models that lead to the emergence of correct finger gaits. We then use the learned policy as a skill primitive within a teleoperation system to collect real-world demonstrations that contain tactile and proprioceptive information. Finally, we train a behavior cloning policy that incorporates tactile sensing and show that it generalizes to nuts and screwdrivers with diverse geometries. Experiments across both tasks show high task progress ratios compared to direct sim-to-real transfer and robust performance even on unseen object shapes and under external perturbations.
\end{abstract}

\vspace{-0.2em}
\section{Introduction}
\vspace{-0.2em}

Reinforcement learning (RL) paired with sim-to-real transfer has recently delivered a number of promising results in dexterous manipulation~\cite{openai2018learning,openai2019solving,handa2023dextreme,chen2023visual,qi2022hand}. Policies trained in massively parallel simulation~\cite{makoviychuk2021isaac} with domain randomization~\cite{peng2018sim} have demonstrated strong robustness and generalization capabilities in the real world.

However, in practice, sim-to-real transfer faces two major limitations. First, due to the complexity of physics simulation, only a limited range of tasks can be accurately modeled. Prior work either relies on specialized techniques for high-fidelity simulation~\cite{narang2022factory,heiden2021disect} or seeks generalization through domain randomization~\cite{lee2020learning,kumar2021rma,openai2018learning,loquercio2021learning}. However, as the task becomes more dynamic, the sim-to-real gap grows~\cite{wang2024lessons}, and simulation alone becomes insufficient. Second, existing sensing modalities have an intrinsic sim-to-real gap. While vision can be partially mitigated through domain randomization~\cite{singh2024dextrah}, tactile sensing remains difficult to approximate reliably. Although some work aims to improve tactile simulation~\cite{wang2022tacto,akinola2025tacsl} or trains policies using alternative proxy representations~\cite{qi2023general,yang2024anyrotate,yin2025learning}, these approaches cannot leverage the full power of tactile sensing. These limitations remain widely viewed as major constraints on the complexity of tasks that can be achieved.

On the other hand, teleoperation and imitation learning~\cite{chi2025diffusion,zhao2023learning} remove the need for simulation entirely. In this setting, policies can learn directly from real-world interactions and sensorimotor signals, which avoids the challenges introduced by sim-to-real transfer. However, teleoperating dexterous hands is challenging because of the intrinsic morphology differences between human and robot hands~\cite{arunachalam2022dexterous,arunachalam2023holo,qin2023anyteleop}. As a result, it is difficult to collect datasets that are large and diverse enough to achieve the desired behavior and generalization.

Motivated by these observations, we introduce \ours, a framework that combines the strengths of both approaches to expand the capability of sim-to-real reinforcement learning under imperfect simulation. The key idea is that \textit{the motion primitives underlying contact-rich dexterous manipulation do not need to be learned from a perfect physics model}. A simplified simulator is sufficient to induce the core rotational behaviors required for these tasks. Once this motion is learned, the resulting policy can be used as a skill primitive to collect real-world demonstrations, from which a new policy can be learned. In this way, sensing modalities and physical interactions that are difficult to simulate can be obtained directly from real-world data, while the fine-grained motions that are hard to teleoperate are provided by the simulation-trained policy. We demonstrate this idea through two tasks, nut-bolt fastening and screwdriving with a multifingered hand. Both tasks are traditionally viewed as requiring complex contact dynamics understanding and tactile sensing. We instead show that effective policies can be learned without relying on high-fidelity simulation.

More specifically, our framework consists of three stages. First, we train reinforcement learning (RL) policies in simulation using a simplified physics model. Instead of modeling the thread structure of the nut and screw, we approximate their interaction with a revolute joint that connects two simple geometric shapes, which allows the policy to efficiently learn rotational behavior. Second, we use this learned skill as a primitive within a teleoperation system to collect real-world demonstrations. The operator controls the arm motion and triggers the finger rotation skill rather than issuing low-level joint commands, which enables efficient collection of tactile data during teleoperated execution. Finally, using the resulting multisensory dataset, we train a behavior cloning policy that coordinates arm and finger motions while leveraging tactile feedback.

We evaluate our framework on two tasks: nut-bolt fastening and screwdriving. Policies trained with simplified dynamics can generate reasonable rotational behavior but cannot complete the tasks. By learning from real-world multisensory demonstrations, our method overcomes these limitations and achieves stable and reliable performance under challenging contact conditions. These results show that complex contact-rich manipulation skills can be bootstrapped from simplified simulators and that real-world tactile feedback is essential. Our framework provides a scalable path toward dexterous manipulation and supports broader deployment of general-purpose robot hands in unstructured environments.

\section{Related Work}

Dexterous manipulation has been a long-standing challenge in robotics~\cite{bicchi2002hands,okamura2000overview}. Early work focused on classical model-based control and analytic grasp planning~\cite{rus1999hand,mordatch2012contact,fearing1986implementing,morgan2022complex,patidar2023hand}. Recent years have seen rapid progress in learning-based approaches, which can be grouped into two primary directions: sim-to-real learning paired with reinforcement learning~\cite{openai2018learning,qi2022hand,chen2023visual} and imitation learning from teleoperation~\cite{lin2024learning,arunachalam2023holo,qin2023anyteleop} or human data~\cite{wang2024dexcap,xu2025dexumi}.

Both directions, however, face notable limitations. Sim-to-real methods benefit from large-scale simulated data and can generalize across diverse objects, yet they remain limited by inaccuracies in modeling complex contact dynamics and sensing, a challenge that becomes more significant as task complexity increases~\cite{wang2024lessons,huang2023dynamic}. Imitation learning benefits from multisensory real-world data, yet collecting high-quality dexterous demonstrations is considerably more difficult than collecting data for simpler end-effectors. Our work seeks to combine the strengths of both approaches. We use large-scale simulation to learn motion primitives while leveraging real-world data to close the dynamics and sensing gaps. Moreover, our skill-based framework enables efficient collection of dexterous real-world data by using the simulation-trained policy itself as a reusable skill primitive.

In the context of nut fastening and screwdriving, there has been recent work combining sim-to-real transfer with teleoperation. For example, Liu et al.~\cite{liu2025dexndm} build a residual model from real-world interactions to compensate for the sim-to-real gap and achieve robust in-hand manipulation. Yin et al.~\cite{yin2025dexteritygen} use simulation-trained policies as stability controllers to enable complex manipulation skills. Both approaches can be integrated with teleoperated arm control to complete these tasks. However, they do not produce autonomous policies that incorporate tactile sensing. Another line of work is Kumar et al.~\cite{kumar2025diffusion}, who demonstrate screwdriver turning by combining learning with trajectory optimization. Noseworthy et al.~\cite{noseworthy2025forge} present an autonomous sim-to-real policy but only show results with a parallel-jaw gripper and do not demonstrate regrasping.

Another way to address the sim-to-real gap is to refine the policies in the real world. For example, Transic~\cite{jiang2024transic} shows that sim-to-real policies can adapt to complex real-world dynamics with only a few human interventions as demonstrations, although the demonstrations are primarily performed with simple end-effectors. In contrast, we apply this idea to dexterous hands. Maddukuri et al.~\cite{maddukuri2025sim} show that co-training with both simulation and real-world data can reduce the gap and improve manipulation performance. RLPD~\cite{zhang2025traversability} and Sparsh-X~\cite{higuera2025tactile} share a similar philosophy of augmenting modalities through human-guided data collection; however, they are not motivated by the possibility of learning skills from imperfect simulations, as their policies are designed to be directly transferable.

\begin{figure*}[t]
    \centering
    \includegraphics[width=\textwidth]{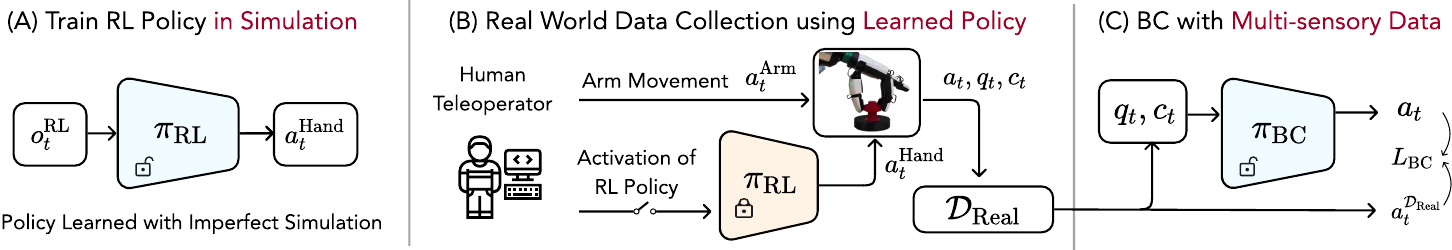}
    \caption{\textbf{An overview of our approach.} We first train a reinforcement learning policy in simulation using a simplified object model, which serves as a motion prior for nut-bolt fastening and screwdriving. We then collect real-world trajectories by using the learned policy as a skill primitive during teleoperation. Finally, we train a behavior cloning policy on the collected data to obtain coordinated behavior between the arm and the fingers.}
    \label{fig:framework}
\end{figure*}

\section{Dexterity from Imperfect Simulation}

An overview of our method is shown in Figure~\ref{fig:framework}. It consists of three stages. First, we train a reinforcement learning (RL) policy in simulation using a simplified object model (Section~\ref{subsec:rl}). The resulting policy learns the desired finger motions but does not experience real-world dynamics and lacks tactile feedback. To address this, we collect real-world trajectories using the learned policy as a skill primitive for teleoperation (Section~\ref{subsec:teleop}). Finally, using this dataset, we train a new multisensory policy using behavior cloning (Section~\ref{subsec:bc}).

\subsection{Training a Reinforcement Learning Policy in Simulation}
\label{subsec:rl}

\paragraphc{Simplified Object Modeling.} Our goal is to design a simulation environment that enables fast training and encourages the emergence of desired finger gaits for rotation. To achieve this, we construct a simplified simulated object (Figure~\ref{fig:sim_objects}) that captures the essence of rotational motion. The object consists of a fixed cylindrical base with a nut or handle attached via a revolute joint. This setup allows the policy to learn rotational motion efficiently without relying on expensive contact-rich simulations. A similar idea was explored in~\cite{lin2024twisting} to model bottle caps using a heuristic friction design. In contrast, we further simplify the model, since we can leverage real-world data to compensate for the resulting dynamics mismatch.

Specifically, for the nut-bolt task, we use a thick triangular shape as the training object (Figure~\ref{fig:sim_objects}A). The extra thickness is used to prevent the policy from learning suboptimal strategies that apply a large force from the bottom. The learned policy also discovers a high-clearance gait that transfers well to diverse real-world nuts such as hexagonal and cube-shaped nuts. For the screwdriver task, where the primary difficulty arises from slippage around the handle, we use spherical primitives to keep the learned behavior conservative (Figure~\ref{fig:sim_objects}B). This observation, that different training shapes lead to different rotational gaits, is also consistent with the findings in~\cite{qi2022hand}. Note that these objects do not need to be visually aligned with real world objects, as they are only used to learn the coarse motions used for real world data collection, as we discussed in Section~\ref{subsec:teleop}.

\begin{figure}[t]
\centering
\includegraphics[width=\linewidth]{figures/sim_object.pdf}
\caption{\textbf{Simplified Object Models.} Each nut or handle is modeled as a rigid body attached to a fixed base through a revolute joint. This abstraction ignores thread-level mechanics while retaining the essential rotational dynamics needed for learning.}
\label{fig:sim_objects}
\end{figure}

\paragraphc{Training Pipeline.} Following~\cite{qi2023general,wang2024lessons}, we first train an oracle policy and then distill it into a sensorimotor policy. The oracle policy $\bm{f}$ is trained with access to an embedding of privileged information~\cite{chen2020learning} $\bm{z}_t$. The sensorimotor policy operates without privileged sensing and instead conditions on a predicted embedding $\hat{\bm{z}}_t = \bm{\phi}(\bm{h}_t)$ inferred from proprioceptive history $\bm{h}_t$ by a prediction module $\bm{\phi}$.

\paragraphc{Privileged Information.} The oracle policy has access to ground-truth environment and object properties, including object attributes (e.g., position, scale, mass, center of mass, friction coefficients), hand pose and finger configurations, and low-level controller parameters. The full set of privileged inputs is documented in the appendix.

\paragraphc{Actions.} At each step, the policy outputs a relative target position. The position command is computed as $\bm{a}_t^{\text{Hand}} = \eta \bm{f}(\bm{o}^{\text{RL}}) + \bm{a}_{t-1}^{\text{Hand}}$, where $\eta$ is the action scale. This command is sent to the robot and converted into torque via a low-level PD controller. Here, $\bm{o}^{\text{RL}}$ contains the robot's proprioceptive state, including joint positions and previous target positions from a sliding window of recent 3 timesteps.

\paragraphc{Reward.} The goal of the policy in simulation is to rotate the simplified object around the revolute joint. The reward consists of a task reward, energy penalties, and stability penalties (time index $t$ omitted for simplicity). Each component includes several terms defined in the appendix: $$r_t = \lambda_{\text{task}} r^{\text{task}} + \lambda_{\text{energy}} r^{\text{energy}}_t + \lambda_{\text{stability}} r^{\text{stability}}_t.$$

\paragraphc{Oracle Policy Training.} We train the oracle policy using proximal policy optimization (PPO)~\cite{schulman2017proximal} with the reward described above. The robot state and privileged information are each encoded with separate MLPs. These embeddings are concatenated and passed through an MLP to produce the final action and value predictions. We train the policy for 1.5$\times$10$^{\text{9}}$ environment steps.

\paragraphc{Sensorimotor Policy Training.} The sensorimotor policy receives proprioceptive states and a latent code $\hat{\bm{z}}_t = \bm{\phi}(\bm{h}_t)$ inferred from a 30-timestep history. We train the policy using DAgger~\cite{ross2011reduction}: at each step, the sensorimotor policy acts in the environment, while the oracle policy provides target actions and ground-truth privileged embeddings. The training objective is
\[
\mathcal{L} = \| a^\text{Hand}_t - \hat{a}^\text{Hand}_t \|_2^2 + \| z_t - \hat{z}_t \|_2^2,
\]
where $\hat{a}^\text{Hand}_t$ denotes the actions produced by the sensorimotor policy. The embedding predictor $\phi$ is optimized using Adam~\cite{kingma2014adam} until convergence.

\paragraphc{Randomization.} We apply domain randomization during training to improve the robustness of the RL policy~\cite{peng2018sim}. Following~\cite{wang2024lessons}, we randomize the nut/handle mass, center of mass, friction coefficient, size, and PD gains, and we also add observation noise. Detailed parameters are provided in the appendix.

\paragraphc{Termination Conditions.} To prevent the policy from getting stuck in unrecoverable states, we terminate an episode when any of the following conditions are met: (1) the distance between the thumb or index finger and the nut/handle exceeds a reset threshold (\SI{7}{\centi\meter} for nuts, \SI{10}{\centi\meter} for handles); (2) the nut or handle remains stagnant over a sliding time window (\SI{3.5}{\second} for nuts, \SI{3}{\second} for handles); or (3) near-zero contact forces are detected for the same duration. These conditions accelerate training by penalizing failure modes such as drifting away from the object or failing to maintain contact.

\begin{figure}[t]
    \centering
    \includegraphics[width=\linewidth]{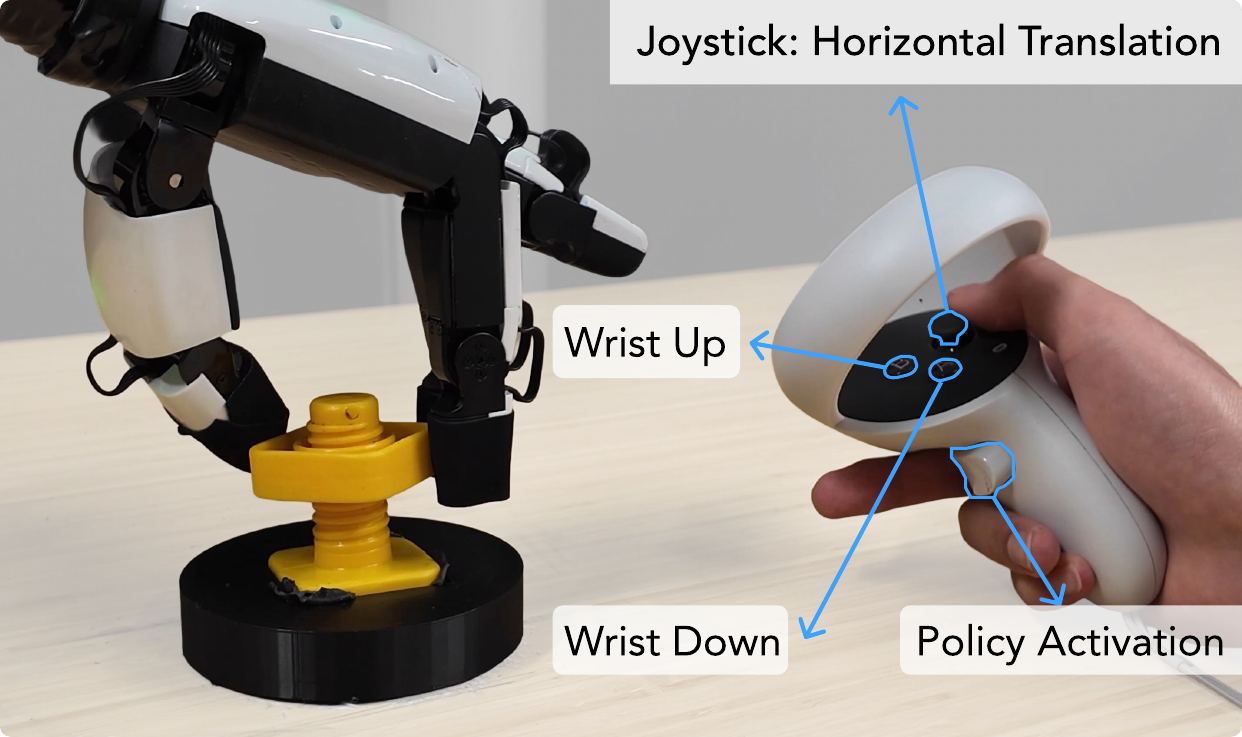}
    \caption{\textbf{Teleoperation Interface.} The human operator controls the wrist position using the VR controller buttons and adjusts yaw and pitch through the joystick. This setup allows the operator to guide the arm motion while relying on the learned finger-rotation skill during data collection.}
    \label{fig:teleop}
\end{figure}

\subsection{Real World Data Collection with Learned Policy}
\label{subsec:teleop}

The policy trained in simulation with the simplified object model learns the desired rotational behavior; however, it inevitably misses key physical dynamics such as thread interactions. These effects are difficult to model but are crucial for reliable real-world fastening. It also lacks tactile information, which is hard to simulate but crucial for fine-grained wrist adjustments. 

To bridge this gap, we introduce a skill-based \emph{assisted teleoperation} for real-world data collection. The core idea is to reuse the simulation-trained policy as a skill primitive for finger motion control. Instead of commanding every joint individually, the human operator controls only the wrist movement and decides when to activate the skill primitive (Figure~\ref{fig:teleop}). Wrist position is specified using the Quest VR\footnote{https://www.meta.com/quest/products/quest-2} controller’s joystick. This approach is inspired by~\cite{lin2024learning}, but we use a much finer-grained skill for data collection.

This framework offers several advantages. First, it delegates complex finger motions to a robust simulation-trained policy that generalizes across different objects, eliminating the need for humans to learn finger coordination under morphological differences. Second, using a joystick for arm control enables precise and intuitive wrist positioning. While arm motion is relatively straightforward to simulate, we intentionally exclude it from the initial simulation training stage, as learning the specific downward progression required for nut-bolt fastening would necessitate simulating thread-level mechanics. Finally, collecting data in the physical environment provides the multisensory observations, particularly tactile signals, that are essential for reliable task completion but remain difficult to approximate reliably in simulation.

Concretely, at each timestep we record two actions: (1) the action generated by the RL policy $\pi_{\rm RL}$, which defaults to the current hand joint positions when the policy is not activated, and (2) the arm action generated by human teleoperation. Formally, we define $\bm{a}_t = [\bm{a}_t^{\text{Hand}}, \bm{a}_t^{\text{Arm}}]$, where $\bm{a}_t^{\text{Hand}} \in \mathbb{R}^{12}$ denotes the hand target joint positions, and $\bm{a}_t^{\text{Arm}} \in \mathbb{R}^{6}$ denotes the arm joint positions. We also record the multisensory observation $\big(\bm{q}_t, \bm{c}_t \big)$, where $\bm{q}_t = [\bm{q}_t^{\text{Hand}}, \bm{q}_t^{\text{Arm}}]$ contains all joint positions, and $\bm{c}_t$ represents the raw tactile signals from all five fingers.

\paragraphc{Tactile Signal.} In this work, we use the XHand’s built-in tactile sensors to capture contact information. Each fingertip is equipped with a pressure-based tactile array comprising 120 sensing elements, each measuring three-axis forces with a minimum detectable force of \SI{0.05}{\newton}. At each timestep, we record the tactile signal as $\mathbf{c}_t \in \mathbb{R}^{5 \times 120 \times 3}$, which includes the signals from all five fingers across three axes.

\subsection{Behavior Cloning with Multisensory Data}
\label{subsec:bc}

With the dataset $\mathcal{D}_{\text{Real}}$ collected using the skill-based \textit{assisted teleoperation}, we can train a behavior cloning (BC) policy $\pi_{\text{BC}}$ using the paired multisensory observations and expert actions. Vision is not used in our work.

\paragraphc{Neural Network Architecture.} We use a feedforward network as the policy. The past $K$ timesteps of observations $(\bm{q}_{t-K+1:t}, \bm{c}_{t-K+1:t})$ are concatenated into a single feature vector. Tactile signals are first flattened and passed through an MLP. The fused feature vector is then processed by an hourglass encoder~\cite{newell2016stacked}, which outputs the action predictions.
We also apply an action chunking strategy~\cite{zhao2023learning,chi2025diffusion}, where the policy predicts a sequence of future actions $\hat{\bm{a}}_{t:t+H}$ rather than a single-step action. We use $K=5$ and $H=16$ unless otherwise noted.

\begin{table*}[t]
\centering
\setlength{\tabcolsep}{3.5pt}
\renewcommand{\arraystretch}{1.05}
\caption{Real-world fastening performance on square, triangular, hexagonal, and cross-shaped nuts. We report progress ratio and rotation time (mean ± standard deviation over 10 trials) for different observation modalities. Tactile sensing and temporal history both improve performance, and their combination yields the highest accuracy and fastest execution. * indicates that only one completely successful trial was recorded, so no standard deviation is reported.}
\resizebox{\linewidth}{!}{
\begin{tabular}{cccccccccc}
\toprule
& &
\multicolumn{2}{c}{Square Nuts} & \multicolumn{2}{c}{Triangular Nuts} & \multicolumn{2}{c}{Hexagonal Nuts} & \multicolumn{2}{c}{Cross-Shaped Nuts} \\
\cmidrule(lr){3-4}
\cmidrule(lr){5-6}
\cmidrule(lr){7-8}
\cmidrule(lr){9-10}
Tactile & Hist. & Prog. Ratio (\%) $\uparrow$ & Time (s) $\downarrow$ & Prog. Ratio (\%) $\uparrow$ & Time (s) $\downarrow$ & Prog. Ratio (\%) $\uparrow$ & Time (s) $\downarrow$ & Prog. Ratio (\%) $\uparrow$ & Time (s) $\downarrow$ \\
\midrule
& & \pms{63.75}{33.05} & \pms{148.76}{30.57} & \pms{30.00}{39.62} & 229.39*  & \pms{75.00}{29.46} & \pms{102.36}{25.00} & \pms{63.75}{33.05} & \pms{101.07}{0.51} \\
\checkmark & & \pms{62.50}{33.85} & \pms{202.21}{59.69} & \pms{66.25}{27.67} & \pms{134.19}{46.72} & \pms{75.00}{16.67} & \pms{205.02}{82.44} & \pms{82.50}{10.54} & \pms{127.17}{47.98} \\ 
& \checkmark & \pms{87.50}{20.41} & \pms{129.88}{48.06} & \pms{80.00}{32.91} & \textbf{\pms{111.47}{55.10}}  & \pms{85.00}{32.17} & \textbf{\pms{68.97}{13.99}} & \textbf{\pms{100.00}{00.00}} & \pms{91.49}{44.89}\\
\midrule
\checkmark & \checkmark & \textbf{\pms{97.50}{7.91}} & \textbf{\pms{124.20}{33.22}} & \textbf{\pms{96.25}{8.44}} & \pms{117.79}{52.13} & \textbf{\pms{95.00}{10.54}} & \pms{75.07}{17.41} & \pms{98.75}{3.95} & \textbf{\pms{84.21}{57.19}} \\
\bottomrule
\end{tabular}
}
\label{tab:nut-bolt}
\end{table*}

\paragraphc{Training.} The policy is trained with supervised learning using the loss
\[
\mathcal{L}_{\text{BC}} =
\sum_{t=1}^{T} 
\sum_{h=0}^{H}
\left\|\, \hat{\bm{a}}_{t+h} - \bm{a}_{t+h} \,\right\|_2^2,
\]
where $\hat{\bm{a}}_{t:t+H}$ is the action chunk predicted by $\pi_{\text{BC}}$, and $\bm{a}_{t:t+H}$ is the corresponding expert sequence. This objective encourages consistent predictions over the full horizon.
We train the policy using Adam~\cite{kingma2014adam} for 200 epochs and normalize observations following~\cite{barreiros2025careful}.

\section{Experiments}

In this section, we first introduce the experiment setup (Section~\ref{subsec:setup}). We then evaluate the performance of our policies on two challenging tasks, nut-bolt fastening (Section~\ref{subsec:nut-bolt}) and screwdriving (Section~\ref{subsec:screwdriver}). We conclude with qualitative experiments that provide additional analysis and design ablations in simulation.

\subsection{Experiment Setup}
\label{subsec:setup}

\paragraphc{Hardware Setup.} Our system consists of a UR5e robot arm (6 DoF) and a 12-DoF XHand. The XHand has five fingers: the thumb and index finger each have 3 DoF, and the remaining fingers have 2 DoF. Only the thumb and index provide abduction/adduction.

\paragraphc{Simulation.} We train our policies in IsaacGym~\cite{makoviychuk2021isaac} using 8{,}192 parallel environments. Each environment contains a simulated XHand and the simplified object model described in Section~\ref{subsec:rl}. The simulation runs at \SI{200}{\hertz}, with control applied at \SI{20}{\hertz}. Each episode lasts up to 800 simulation steps (\SI{40}{\second}).

\paragraphc{Object Set.} For the nut-bolt task, we train on triangular nuts. For the screwdriver task, we approximate handles by octagon- and dodecagon-shaped nuts. This multi-geometry training in simulation helps the policy generalize to diverse real-world shapes.

\paragraphc{Metrics.} In simulation, we report the episode reward and episode length during training. In real-world evaluation, we measure the \emph{progress ratio}, defined as the number of successful rotations divided by the total number of rotations required for full fastening. We also report the \emph{time} for trials that fully complete the fastening process (progress ratio = 100), defined as the time needed to fully fasten the nut or fully tighten the screw. Note that some baseline methods do not achieve a single successful fastening or screwing attempt across ten trials. In such cases, we cannot report the standard deviation or the completion-time metric.

\subsection{Nut-Bolt Experiments}
\label{subsec:nut-bolt}

We first evaluate the system on the nut-bolt task. This task requires the fingers to establish correct contact patterns, sense progress through tactile feedback, and adjust the arm position accordingly. We choose this task because nut-bolt interactions are difficult to simulate efficiently, and completing the task relies heavily on tactile sensing, making it a strong testbed for our method.

We note that direct sim-to-real transfer can rotate the nut, but it cannot drive the nut downward because the arm does not move. Since thread interactions are not simulated, the nut remains at the same height even after completing full revolutions in simulation.

\paragraphc{Setting.} In simulation, we use the thick triangular nut shown in Figure~\ref{fig:sim_objects} (left). Training with this shape produces high-clearance gaits that transfer well and can rotate both square and triangular nuts in the real world.

During skill-based assisted teleoperation, we collect 50 trajectories each for the square and triangular nuts. Each trajectory lasts about 80 seconds. We then train a behavior cloning policy using the combined dataset. We evaluate performance on four types of nuts, which include square and triangular nuts as well as two unseen shapes, namely hexagonal nuts and cross-shaped nuts. Our main results are shown in Table~\ref{tab:nut-bolt}.

\paragraphc{Observation History.} We first study the effect of providing a short temporal history in the observation. Adding history significantly improves progress ratio and reduces execution time across all modalities and nut geometries. Temporal cues help the policy track fine-grained rotational progress and distinguish local geometric features. The benefit is especially clear for non-tactile policies, where history stabilizes performance and narrows much of the gap to tactile-based policies on easier geometries. When combined with tactile sensing, history yields the strongest overall performance, achieving the highest accuracy and fastest completion times across all nut types, including unseen shapes.

\paragraphc{Tactile Information.} Across all nut geometries, adding tactile input generally improves progress ratio. The effect is most pronounced on challenging shapes such as triangular and cross-shaped nuts, where progress ratios rise from roughly 30 to 65 percent for triangular nuts and from about 60 to 80 percent for cross-shaped nuts. This underscores the importance of tactile feedback for maintaining stable contact and detecting effective rotation. We also observe that certain non-tactile settings achieve high performance, even on unseen shapes, particularly on well-constrained geometries such as hexagonal and cross-shaped nuts. In these cases, the geometry provides strong passive guidance, and the finger gait learned in simulation is often sufficient to maintain contact without relying on tactile cues.

\paragraphc{Failure Modes.} We observe two main failure modes. First, the policy without observation history struggles to infer object shape from single-step proprioception. As a result, it fails to generalize across nut geometries, since different nuts require different rotational gaits. The policy cannot adjust its gait because the limited sensing information prevents it from identifying the correct motion pattern. Second, non-tactile policies frequently drift into unstable contact states and lose alignment with the bolt. Once misalignment occurs, the nut cannot sustain continuous rotation. In contrast, tactile policies can recover by adjusting wrist orientation or applying corrective downward force to re-establish contact.

\subsection{Screwdriving Experiments}
\label{subsec:screwdriver}

\begin{table}[t]
\centering
\setlength{\tabcolsep}{4pt}
\renewcommand{\arraystretch}{1.1}
\caption{\textbf{Real-world screwdriving performance}. We report progress ratio and rotation time (mean ± standard deviation over 10 trials) for direct sim-to-real, expert replay, and our behavior cloning (BC) ablations. Tactile sensing and temporal history each improve performance, and their combination achieves the highest progress ratio and fastest execution. Here * indicates that the policy never fully completed the task, so no rotation time is reported.}
\resizebox{\linewidth}{!}{
\begin{tabular}{ccccc}
\toprule
& & & \multicolumn{2}{c}{Screwdriver Task} \\
\cmidrule(lr){4-5}
Method & Tactile & Hist. & Prog. Ratio (\%) $\uparrow$ & Time (s) $\downarrow$ \\
\midrule
Direct Sim2Real & & & \pms{41.60}{26.21} & N.A.* \\
Expert Replay & & & \pms{50.80}{19.27} & N.A.* \\
\midrule
\multirow{4}{*}{Ours} & & & \pms{69.20}{35.25} & \pms{266.33}{91.60} \\
& \checkmark & & \pms{67.63}{35.65} & \pms{264.06}{76.57} \\
&  & \checkmark & \pms{87.50}{18.61} & \pms{195.15}{44.83} \\
& \checkmark & \checkmark & \textbf{\pms{95.00}{13.24}} & \textbf{\pms{187.87}{24.87}} \\
\bottomrule
\end{tabular}
}
\label{tab:screwdriver}
\end{table}

We also demonstrate that our method extends to the more challenging screwdriving task. Compared to nut-bolt fastening, screwdriving is inherently less stable: the shaft is not kinematically constrained along the screw axis, and even small tilts or misalignments can lead to slipping or loss of contact. As a result, the task requires more fine-grained control to maintain continuous rotation. The complex interaction between the screwdriver and the screw is also difficult to simulate accurately, which is why we include this task in our study.

\paragraphc{Setting.} In simulation, we use a mix of octagonal and dodecagonal handles as the training objects, as shown in Figure~\ref{fig:sim_objects} (right). This curated set encourages the learned rotational gaits to remain conservative in clearance and maintain stability. In the real world, we collect 72 trajectories, each lasting between 120 and 180 seconds.

\paragraphc{Sim-to-Real Policy.} Unlike the nut-bolt experiments, the screwdriving task can still make progress even when the wrist does not move, since downward motion is not required. We first evaluate a direct sim-to-real policy that uses only proprioception. The results are shown in Table~\ref{tab:screwdriver}. The direct sim-to-real policy achieves a 41.60\% progress ratio, indicating that it can produce meaningful behavior, which is a prerequisite for collecting expert trajectories. However, because it inevitably makes mistakes, it never completes the task even once, and therefore we cannot report statistics for completion time. In contrast, during data collection, the human operator can adjust the wrist position to recover from such mistakes.

As a result, when we replay the expert data from the data we collected, it can achieve higher success rate to be 50.80\%. However, because it cannot adapt changes during deployment. It also fails at completely finishing the task. Next, we show the results of our policies.

\begin{figure}
    \centering
    \includegraphics[width=\linewidth]{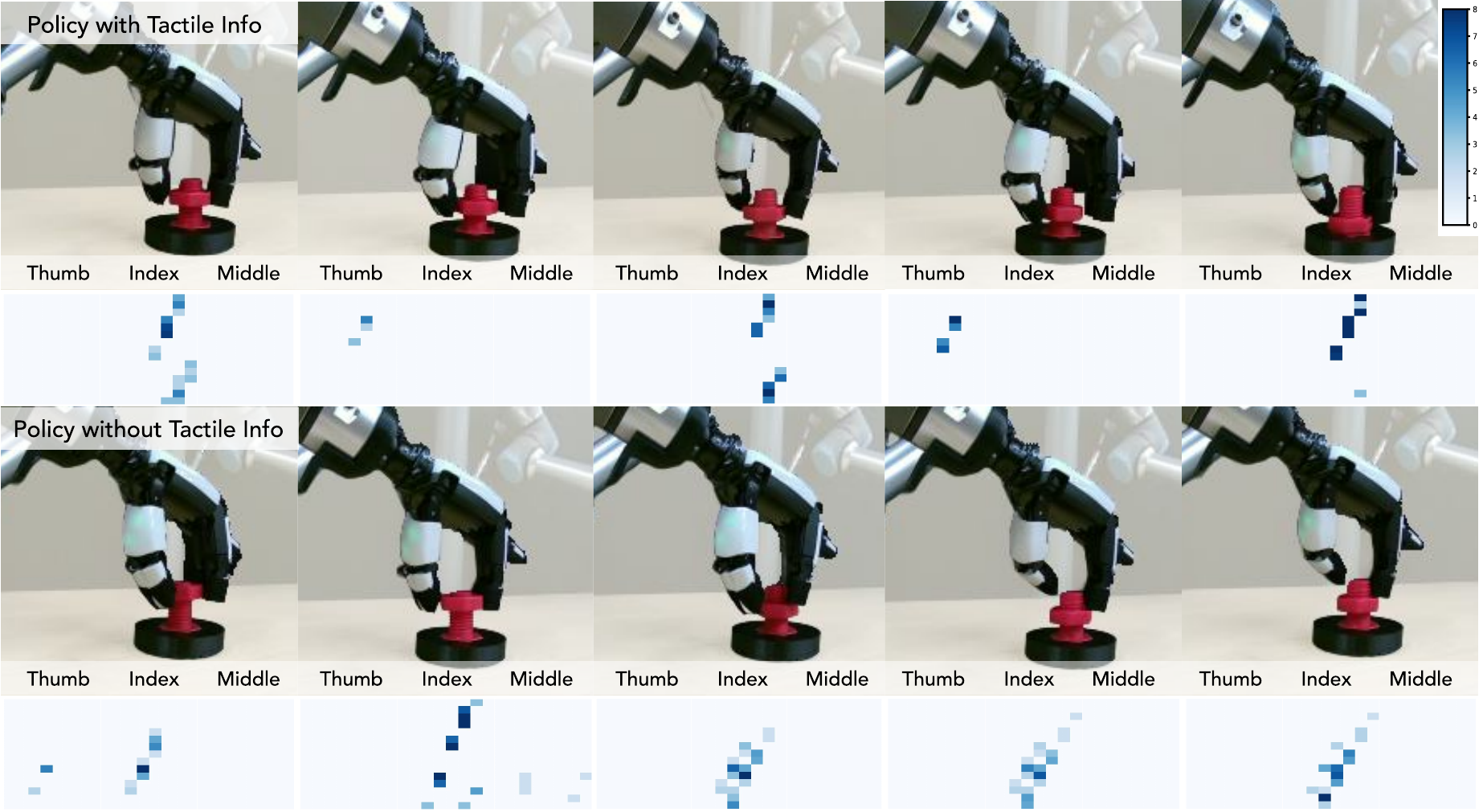}
    \caption{\textbf{Top}: The policy with tactile information maintains a consistent alternating pattern of thumb and index finger contact, which supports stable engagement as the nut is rotated downward. \textbf{Bottom}: The policy without tactile information does not maintain a clear contact pattern. This leads to unsuccessful engagement and prevents proper downward wrist motion. The resulting pattern reflects the index finger pressing against the bolt after losing stable contact.}
    \label{fig:tactile} 
\end{figure}

\paragraphc{Main Results.} Our behavioral cloning policies show clear improvements over the direct sim-to-real and expert-replay baselines. Adding tactile sensing or temporal history individually improves progress ratio. Combining both modalities gives the strongest performance. 

The baseline behavior cloning model already achieves a 69.20\% progress ratio, which substantially outperforms expert replay. This phenomenon, where a behavior cloning policy surpasses the policy that generated the data, is consistent with filtered behavior cloning~\cite{kumar2022should}. In this approach, only successful trials are used for training. A similar effect has also been reported in~\cite{wang2024lessons}.

Using history alone produces a comparable improvement of 67.63\%. This indicates that temporal information helps the policy track rotational progress and recover from partial failures. When history is not used, tactile information provides only limited benefit. However, once history is included, the two modalities become complementary. The progress ratio increases to 95.00\%, and the average rotation time decreases substantially. These results show that tactile feedback and temporal history work together to produce stable, consistent, and efficient screwdriving behaviors.

\paragraphc{Failure Modes.} We observe that open-loop baselines frequently fail due to gradual handle slipping and accumulated orientation drift. Without feedback, small misalignments grow over time. Among behavioral cloning methods, the policy trained without observation history struggles to stabilize the screwdriver since it lacks the temporal information needed to infer handle pose and orientation. Non-tactile policies fail to detect subtle torque imbalances and often lose stable contact under slight perturbations. With both tactile feedback and temporal history, BC with tactile and history compensates for these effects by adjusting wrist orientation and applying appropriate forces.

\subsection{Qualitative Experiments}
\label{sec:tactile}

\begin{figure}[t]
    \centering
    \includegraphics[width=\linewidth]{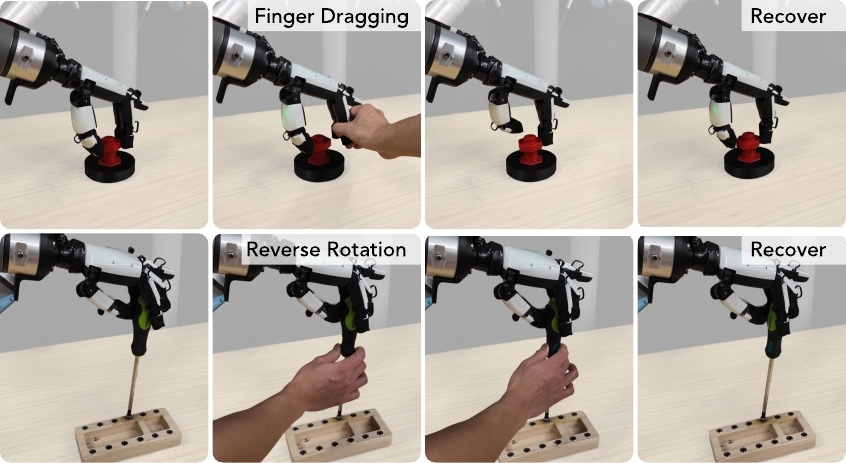}
    \caption{\textbf{Top row:} The policy recovers back to the nut-bolt fastening motion when the fingers are dragged by an external force. \textbf{Bottom row:} The policy recovers back to the screwdriving motion when the screwdriver is rotated counterclockwise during the clockwise rotation by the policy.}
    \label{fig:robust}
\end{figure}

\paragraphc{Out-of-distribution Robustness.} We study the robustness of the learned policy under external perturbations that are not encountered during training. These disturbances include (1) dragging the fingers away from the object or rotating the screwdriver in the opposite direction of the intended direction. In Figure~\ref{fig:robust}, we show that despite these perturbations, the policy consistently recovers to stable fastening behavior. Specifically, it recovers from reverse rotation of the fastening object by re-establishing contact and restoring the correct rotational direction. When finger contact is disrupted or temporarily blocked, the policy repositions the fingers and wrist to regain stable engagement.

\paragraphc{Tactile Visualization.} Tactile signals provide structured spatiotemporal patterns that the policy uses to infer contact phases. As shown in Fig.~\ref{fig:tactile}, stable activation signatures emerge when the nut or screwdriver handle is correctly engaged. The policy learns to preserve these patterns by adjusting wrist orientation and contact force, effectively using tactile feedback as a local reference for alignment. When these patterns deviate, corrective actions such as re-engagement or downward pressing are triggered.

\subsection{Simulation Ablations}
\label{sec:sim_ablation}

\begin{figure}[t]
    \centering
    \includegraphics[width=\linewidth]{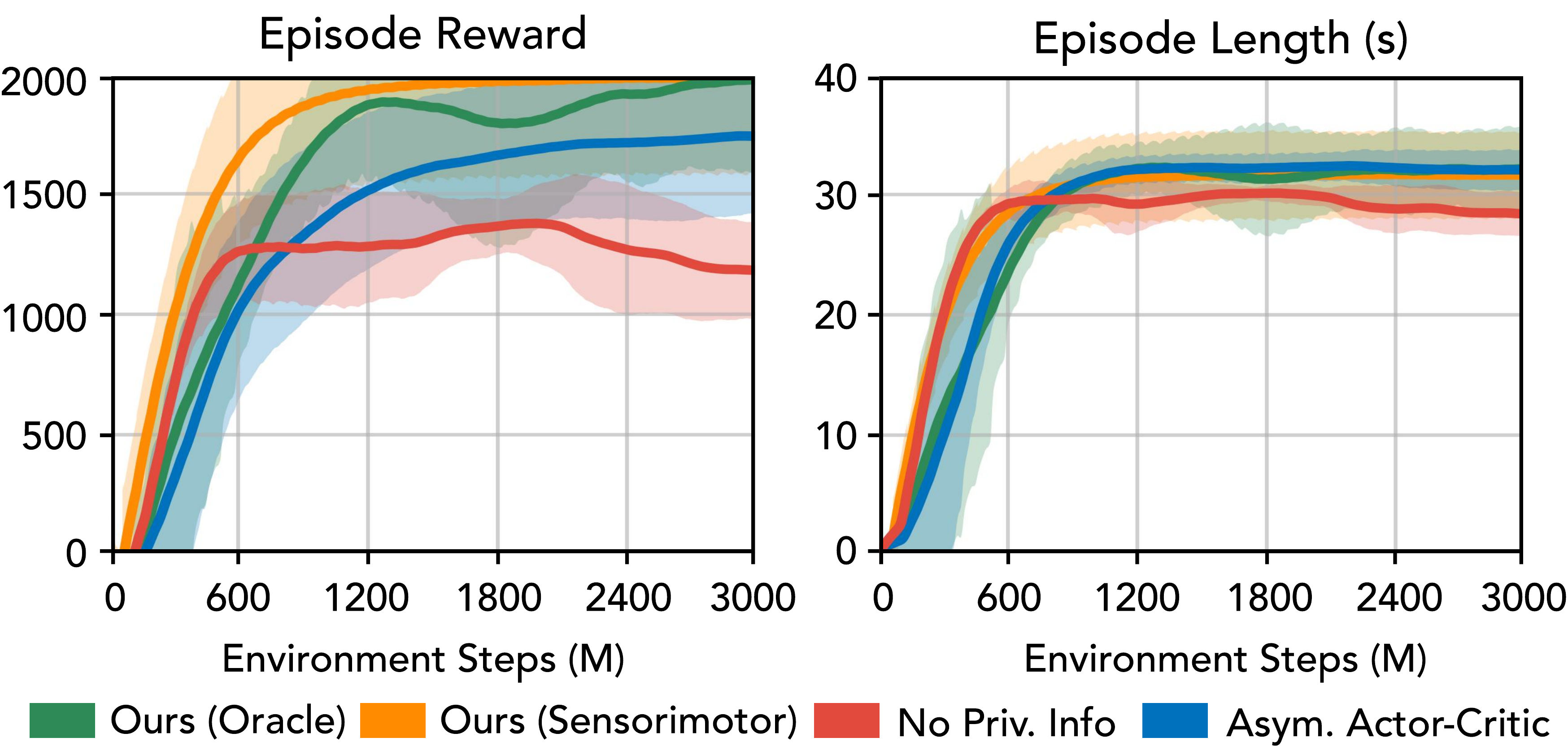}
    \caption{Simulation ablations of screwdriving policy training. We compare our privileged-information oracle policy, its sensorimotor policy, an asymmetric actor–critic variant, and a policy trained without privileged information. Providing privileged information during training leads to significantly higher reward and more stable episode lengths. Each curve shows the mean and standard deviation over 5 seeds.}
    \label{fig:sim_ablation}
\end{figure}

In simulation, we verify the design choices used to train the screwdriving policies. We evaluate our two-stage training procedure against two alternatives: training without any privileged information, and using an asymmetric actor-critic~\cite{pinto2017asymmetric} architecture where the critic has access to full tactile information while the actor does not. The actor in this setting is directly deployable in the real world. The results are shown in Figure~\ref{fig:sim_ablation}.

We compare episode reward and episode length across these training strategies. When both the actor and critic have access to privileged information, our method (Ours, Oracle) achieves the highest performance. Removing privileged information from the actor, as in the asymmetric actor-critic variant, leads to a noticeable drop in performance. Removing privileged information entirely results in a further decline. These results show that privileged information plays an important role during policy learning. We also observe that a sensorimotor policy can approach similar performance when proprioceptive history is provided as input. We also experimented with training the asymmetric and non-privileged models for longer horizons but have not observed further improvement. This suggests that the performance gap is not due to insufficient training but instead reflects the importance of privileged information during learning.

\section{Conclusion and Future Work}

We present a framework for learning dexterous manipulation skills for contact-rich tasks using imperfect simulation. The approach first learns transferable rotational skills through reinforcement learning with simplified object modeling. It then uses these skills for skill-based teleoperation to collect real-world trajectories, and finally incorporates tactile feedback and learns a sensorimotor policy through behavior cloning. Experiments on nut-bolt fastening and screwdriver usage show that simulation alone cannot capture the complex dynamics required for reliable task execution. However, when behavior cloning is combined with tactile sensing and temporal history, the resulting policies become robust and reliable across diverse and unseen object geometries. This staged pipeline provides a practical and scalable solution for contact-rich manipulation and highlights the value of tactile sensing and skill-based teleoperation as effective bridges between simulation and real-world deployment.

\paragraphc{Limitations.} Although our skill-based teleoperation reduces the burden on the human operator, it remains a constraint for scalable data acquisition. Fully autonomous data collection or learning skill-level guidance from human videos would further improve efficiency. In our current tasks, the nut is already installed and the screwdriver is already inserted. Extending the approach to fully long-horizon assembly will require vision sensing and possibly high-accuracy force-torque sensing. A broader evaluation across more contact-rich manipulation tasks is also needed to assess generality.

\section*{Acknowledgment}
This work is supported in part by the program ``Design of Robustly Implementable Autonomous and Intelligent Machines (TIAMAT)", Defense Advanced Research Projects Agency award number HR00112490425. We thank Mengda Xu for his valuable feedback.

\bibliographystyle{IEEEtran}
\bibliography{IEEEabrv,references}

\clearpage

\section*{Appendix}

\subsection{Privileged Information for Oracle Policy}
\label{appendix:priv}

\paragraphc{Nut-Bolt Task.} The oracle policy receives privileged information about object properties, fingertip states, nut-specific dynamics, hand states, and PD controller gains. The full list of privileged inputs is provided in Table~\ref{tab:priv_info_boltnut}.

\paragraphc{Screwdriver Task.} 
The screwdriver task uses a subset of the privileged information from the nut-bolt task. Specifically, it excludes hand base position, hand orientation, hand joint positions, and PD controller gains. All other privileged inputs remain the same.

\subsection{Reward}
Our reward function is a weighted combination of task rewards, energy penalties, and stability penalties:

\renewcommand{\labelitemii}{$\circ$}
\begin{itemize}[leftmargin=10pt]
    \item \textbf{Task Rewards} encourage successful rotation:
    \begin{itemize}[leftmargin=15pt]
        \item \textbf{Rotation Rewards:} 
        $r^{\text{rot}}_t = \text{clip}(\omega_t,\omega_{\min},\omega_{\max})$. It encourages positive angular velocity $\omega$ along the fastening axis, clipped to $[-4.0, 4.0]$ rad/s.
        \item \textbf{Proximity Rewards:} 
        $r^{\text{prox}}_t = \max(0,\,1 - d_t/d_{\text{thresh}})$. It encourages fingers to remain close to the object, where $d_t$ is the mean distance from thumb and index finger to the nut/handle.
    \end{itemize}

    \item \textbf{Energy Penalties} discourage inefficient motions:
    \begin{itemize}[leftmargin=15pt]
        \item \textbf{Torque Penalty:} $r^{\text{torque}}_t = -\|\tau_t\|^2$ penalizes large joint torques.
        \item \textbf{Work Penalty:} $r^{\text{work}}_t = -(|\tau_t|^\top |\dot{q}_t|)^2$ penalizes excessive joint power.
    \end{itemize}

    \item \textbf{Stability penalties} maintain stable behavior:
    \begin{itemize}[leftmargin=15pt]
        \item \textbf{Pose Difference Penalty:} $r^{\text{pose}}_t = -\|q_t - q^0\|^2$ penalizes deviations from the initial finger configuration (thumb joints masked).
        \item \textbf{Large Rotation Penalty:} $r^{\text{rp}}_t = -\max(0, \omega_t - \omega_{\text{thresh}})$ penalizes excessive angular velocity above threshold $\omega_{\text{thresh}}$ (10.0 rad/s for nut-bolt, curriculum from 7.5 to 15.0 rad/s for screwdriver).
    \end{itemize}
\end{itemize}

We sum the above rewards with weights listed in Table~\ref{tab:reward_weights}.

\begin{table}[!t]
\caption{Privileged information for nut-bolt task. The screwdriver task uses a subset of these features (excludes hand pose and PD gains).}
\centering
\renewcommand{\arraystretch}{1.15}
\begin{tabular}{lc}
\toprule
Privileged Information & Dimension \\
\midrule
Object position & 3 \\
Object scale & 1 \\
Object mass & 1 \\
Object friction coefficient & 1 \\
Object center of mass & 3 \\
Object orientation (quaternion) & 4 \\
Object linear velocity & 3 \\
Object angular velocity & 3 \\
Object restitution & 1 \\
Fingertip positions (2 fingers) & 6 \\
Fingertip orientations (2 fingers) & 8 \\
Fingertip linear velocities (2 fingers) & 6 \\
Fingertip angular velocities (2 fingers) & 6 \\
Nut contact indicator & 1 \\
Nut position & 3 \\
Nut joint velocity & 1 \\
Nut joint position & 1 \\
Screw joint friction & 1 \\
Hand scale & 1 \\
Hand position & 3 \\
Hand orientation (quaternion) & 4 \\
Hand joint positions & 12 \\
PD controller gains ($k_p$) & 12 \\
PD controller gains ($k_d$) & 12 \\
\midrule
\textbf{Total} & \textbf{97} \\
\bottomrule
\end{tabular}
\label{tab:priv_info_boltnut}
\end{table}

\begin{table}[!t]
\caption{Reward function hyper-parameters for nut-bolt and screwdriver tasks.}
\centering
\renewcommand{\arraystretch}{1.15}
\begin{tabular}{lcc}
\toprule
Reward Component & nut-bolt & Screwdriver \\
\midrule
\multicolumn{3}{l}{\textit{Task Rewards}} \\
$\lambda_{\text{rotate}}$ & 6.0 & 2.5 \\
$\lambda_{\text{proximity}}$ & 2.0 & 2.0 \\
\midrule
\multicolumn{3}{l}{\textit{Energy Penalties}} \\
$\lambda_{\text{torque}}$ & -0.1 & -3.0 \\
$\lambda_{\text{work}}$ & -0.01 & -0.01 \\
\midrule
\multicolumn{3}{l}{\textit{Stability Penalties}} \\
$\lambda_{\text{pose}}$ & -0.5 & -0.1 \\
$\lambda_{\text{rotate-penalty}}$ & -0.3 & -0.3 \\
$\lambda_{\text{pc-z}}$ & -1.0 & -1.0 \\
\bottomrule
\end{tabular}
\label{tab:reward_weights}
\end{table}

\begin{table}[!t]
\caption{\textbf{Domain Randomization Parameters.} Object scale is discretely sampled from the specified set and multiplied by the base scale. Mass, center of mass, friction, restitution, and PD controller gains are uniformly sampled at environment initialization. Observation and action noise are sampled i.i.d. from Gaussian distributions at each timestep. Following~\cite{openai2018learning}, we apply a random disturbance force with magnitude $2.0m$ (where $m$ is object mass) with probability 0.25 at each timestep.}
\centering
\renewcommand{\arraystretch}{1.15}
\begin{tabular}{ll}
\toprule
Parameter & Range \\
\midrule
Object Scale (nut-bolt) & $\times$[0.95, 1.05] \\
Object Scale (Screwdriver) & $\times$[0.85, 1.25] \\
Mass & [0.04, 0.06] \SI{}{\kilo\gram} \\
Center of Mass & [-0.001, 0.001] \SI{}{\meter} \\
Coefficient of Friction & [0.5, 8.0] \\
Object Restitution & [0.0, 1.0] \\
PD Controller Stiffness ($k_p$) & [2.7, 3.3] \\
PD Controller Damping ($k_d$) & [0.009, 0.011] \\
Observation Noise (rotation) & $\mathcal{N}(0, 0.01)$ (rad) \\
Observation Noise (translation) & $\mathcal{N}(0, 0.005)$ (m) \\
Action Noise (rotation) & $\mathcal{N}(0, 0.01)$ (rad) \\
Action Noise (translation) & $\mathcal{N}(0, 0.005)$ (m) \\
External Force Scale & $2.0m$ \\
External Force Probability & 0.25 per timestep \\
\bottomrule
\end{tabular}
\label{tab:supp:randomization_range}
\end{table}

\subsection{Training Hyperparameters} The inputs to our oracle policy contains the proprioceptive observations consist of $q_t$ (joint positions over the last 3 timesteps) and $a_{t-1}$ (previous joint targets over the last 3 timesteps). The privileged information includes $p_t$ (5 fingertip positions), $w_t$ (object state with 3D position, quaternion orientation, and angular velocity), and additional features detailed in Table~\ref{tab:priv_info_boltnut}. We follow the domain randomization parameters in Table~\ref{tab:supp:randomization_range}.

We train our oracle policy with PPO, and the training hyperparameters are shown in Table~\ref{tab:oracle_hyperparams}. Specifically, we train with 8,192 parallel environments. Each environment gathers 12 steps of data to train in each epoch of PPO. The data is split into minibatches of size 16,384 and optimized with PPO loss. $\gamma$ and $\lambda$ are used for computing generalized advantage estimate (GAE) returns. We use the Adam optimizer to train PPO and adopt gradient clipping to stabilize training. We train 1.5 billion environment steps in total, which takes less than one day on a single GPU. We train our sensorimotor policy with on-policy behavioral cloning, and the training hyperparameters are shown in Table~\ref{tab:student_hyperparams}.

\begin{table}[!t]
\caption{Hyperparameters for training the oracle policy.}
\centering
\renewcommand{\arraystretch}{1.15}
\begin{tabular}{lc}
\toprule
Hyperparameter & Value \\
\midrule
\# environments & 8192 \\
\# steps & 12 \\
\# minibatch size & 16384 \\
\# Environment steps &  3$\times$10$^{9}$  \\
discount factor ($\gamma$) & 0.99 \\
GAE ($\lambda$) & 0.95 \\
learning rate & 5e-3 \\
clip range & 0.2 \\
entropy coefficient & 0.0 \\
kl threshold & 0.02 \\
max gradient norm & 1.0 \\
\bottomrule
\end{tabular}
\label{tab:oracle_hyperparams}
\end{table}

\begin{table}[!t]
\caption{Hyperparameters for training the sensorimotor policy in simulation.}
\centering
\renewcommand{\arraystretch}{1.15}
\begin{tabular}{lc}
\toprule
Hyperparameter & Value \\
\midrule
\# environments & 48 \\
\# steps & 512 \\
\# minibatches & 4096 \\
learning rate & 1e-3 \\
\bottomrule
\end{tabular}
\label{tab:student_hyperparams}
\end{table}

\end{document}